# FCB-SwinV2 Transformer for Polyp Segmentation

Kerr Fitzgerald[1], Bogdan Matuszewski[1]

[1] University of Central Lancashire, Preston, United Kingdom

**Abstract.** Polyp segmentation within colonoscopy video frames using deep learning models has the potential to automate the workflow of clinicians. This could help improve the early detection rate and characterization of polyps which could progress to colorectal cancer. Recent state-of-the-art deep learning polyp segmentation models have combined the outputs of Fully Convolutional Network architectures and Transformer Network architectures which work in parallel. In this paper we propose modifications to the current state-of-the-art polyp segmentation model FCBFormer. The transformer architecture of the FCBFormer is replaced with a SwinV2 Transformer-UNET and minor changes to the Fully Convolutional Network architecture are made to create the FCB-SwinV2 Transformer. The performance of the FCB-SwinV2 Transformer is evaluated on the popular colonoscopy segmentation benchmarking datasets Kvasir-SEG and CVC-ClinicDB. Generalizability tests are also conducted. The FCB-SwinV2 Transformer is able to consistently achieve higher mDice scores across all tests conducted and therefore represents new state-of-the-art performance. Issues found with how colonoscopy segmentation model performance is evaluated within literature are also reported and discussed. One of the most important issues identified is that when evaluating performance on the CVC-ClinicDB dataset it would be preferable to ensure no data leakage from video sequences occurs during the training/validation/test data partition.

## 1  Introduction

Colorectal cancer often arises from small benign polyps which progress over time to become malignant. This form of cancer accounts for a large proportion of cancer deaths worldwide. Polyp classification, detection and segmentation using deep learning models has the potential to automate the workflow of highly skilled clinicians and improve the early detection rate and characterization of polyps which would in turn help reduce mortality rates. Many deep learning systems based on Fully Convolutional Networks (FCNs) [1] [2] [3] [4] [5] and Transformer Networks (TNs) [6] [7] have been applied to polyp segmentation. Recently efforts to combine both types of architectures for medical image segmentation have been made [8] [9]. The current state-of-the-art model for colonoscopy segmentation is named the Fully Convolutional Branch-TransFormer (FCBFormer) [9]. The architecture of the FCBFormer combines the benefits of both TNs and FCNs by running a model of each type in parallel and combining the outputs which are then passed onto a prediction head for processing.

In this paper we propose modifications to the current state-of-the-art polyp segmentation model FCBFormer. We replace the Transformer Branch (TB) with a SwinV2 Transformer-UNET and make minor modifications to the Fully Convolutional Branch (FCB) to create the FCB-SwinV2 Transformer. We evaluate the performance of the FCB-SwinV2 Transformer on the popular colonoscopy segmentation benchmarking datasets Kvasir-SEG [10] and CVC-ClinicDB [11] and achieve new state-of-the-art performance. Issues found with how colonoscopy segmentation model performance is evaluated within the literature are also reported and discussed.

## 2 FCB-SwinV2 Transformer Architecture

The TB of the original FCBFormer is replaced in this work by a SwinV2-UNET architecture. The motivation for this arises due to the fact that models featuring SwinV2 Transformer [12] architectures are currently competitive on numerous semantic segmentation benchmarks including ADE20K [13].

The SwinV2 Transformer [12] was developed to tackle issues with training stability and resolution gaps between pre-training and fine-tuning that arose with the original Swin Transformer [14]. In the original Swin Transformer training instability was caused by large activation output amplitude discrepancies in different layers of the network. To remedy this problem the authors of the SwinV2 Transformer developed an approach called 'residual post normalization'. In the residual post normalization approach the output of residual blocks is normalized before merging into the main branch of the network. This was shown by the SwinV2 creators to cause much milder activation amplitudes than in the original pre-normalization configuration. The authors of the SwinV2 Transformer also replaced the original dot product attention mechanism due to the finding that learnt attention maps of some blocks were frequently dominated by only a few pixel pairs. The dot product attention mechanism was therefore replaced by a mechanism which computes the attention logit of a pixel pair using a scaled cosine function. Since the cosine function is naturally normalized this helps the network become more insensitive to the amplitude of activations.

The SwinV2 Transformer network [12] first partitions an RGB input image into non-overlapping patches. Each patch is a concatenation of the RGB pixel values which are subsequently passed to a linear embedding layer which projects them to have an arbitrary channel dimension. Patch merging layers then concatenate neighboring patches before passing them through a linear layer. This reduces the number of patches by a factor of 2 and increases the channel dimension by a factor of 2. The output of the patch merging layer is then passed through a number of successive SwinV2 transformer blocks. The successive SwinV2 transformer blocks are shown in Figure 1. These blocks consist of either window based multi-head self-attention (W-MSA) (first block) or shifted window partition multi-head self-attention (SW-MSA) (second block) layers, followed by an MLP with GELU [15] activation layer and stages of layer normalization (LN). Residual connections are also present within the block.

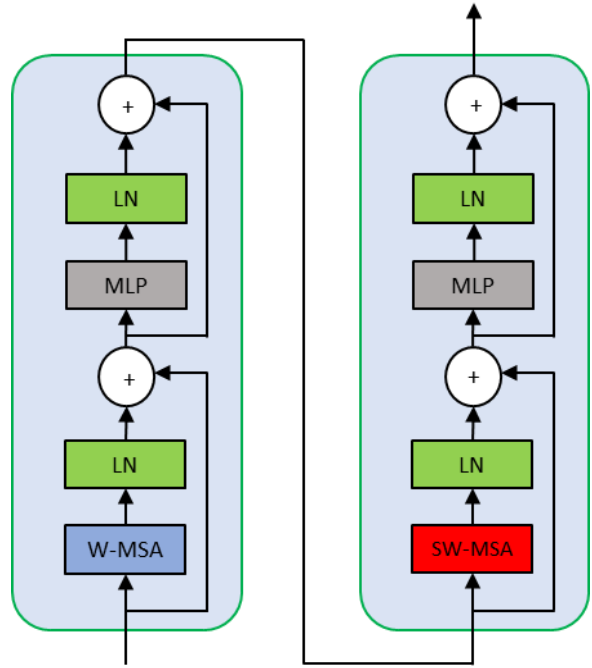

*Figure 1: Two successive SwinV2 Transformer Blocks [12]. The residual post normalization configuration ensures layer normalization is conducted after attention mechanism layers and MLP layers.*

The SwinV2-UNET style system used in this work was based on a system [16] which used a SwinV2 encoder [12] [17] with decoder blocks composed of SCSE modules [18] [19] and standard convolution modules. The SCSE module is composed of a 'Spatial Squeeze and Channel Excitation' (CSE) module and a 'Channel Squeeze and Spatial Excitation' (SSE) module. The CSE module takes an input tensor and reduces the spatial dimension using global average pooling. The resulting tensor is passed through convolutional and activation layers before element wise multiplication with the original input tensor. This results in an output tensor with adaptively re-weighted channel values. The SSE module takes the input tensor and reduces the channel dimension using convolution. The resulting tensor is passed through an activation layer before element wise multiplication with the original input tensor. This results in an output tensor with adaptively re-weighted spatial features. The SCSE module combines the outputs of the CSE and SSE modules using element wise summation, therefore maximizing information propagation through the network at a pixel and channel level simultaneously. The structure of the decoder block and CSCE module is shown in Figure 2.

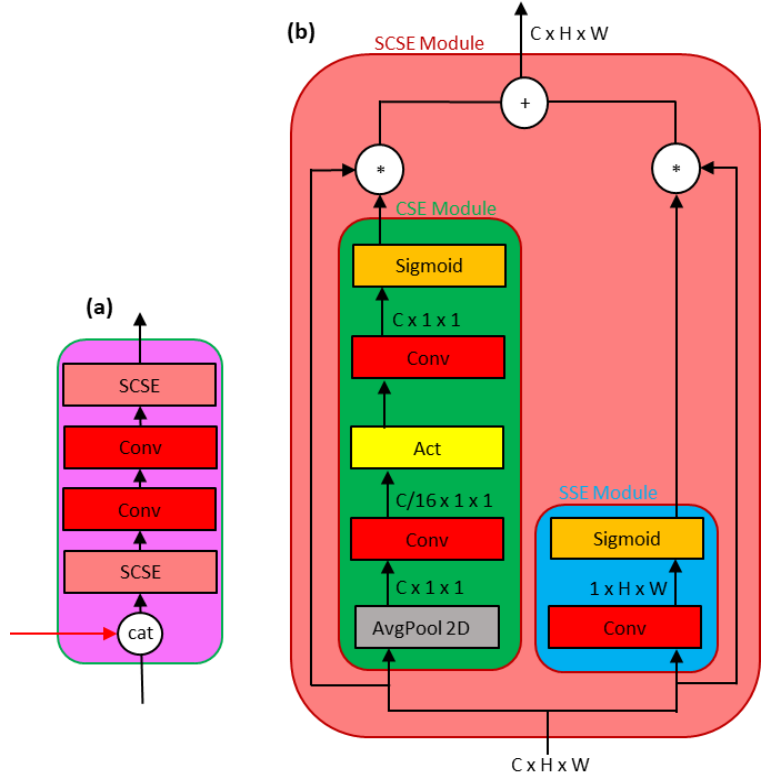

Figure 2: (a) The decoder block [16] uses channel wise concatenation to combine previous decoder layer output with encoder skip connection output. It is necessary after the first decoder layer to use interpolation to scale the spatial dimensions of previous decoder layer output to match the spatial dimensions of the encoder skip connection output. (b) The structure of the SCSE module which combines the output of the CSE and SSE modules.

The resolution of input images into the SwinV2 encoder system (and overall FCB-SwinV2 Transformer system) used in this work is 384x384 due to the availability of ImageNet [20] pre-trained SwinV2 encoder systems available within the PyTorch Image Model library [17]. The SwinV2-UNET system used as the TB is displayed in Figure 3.

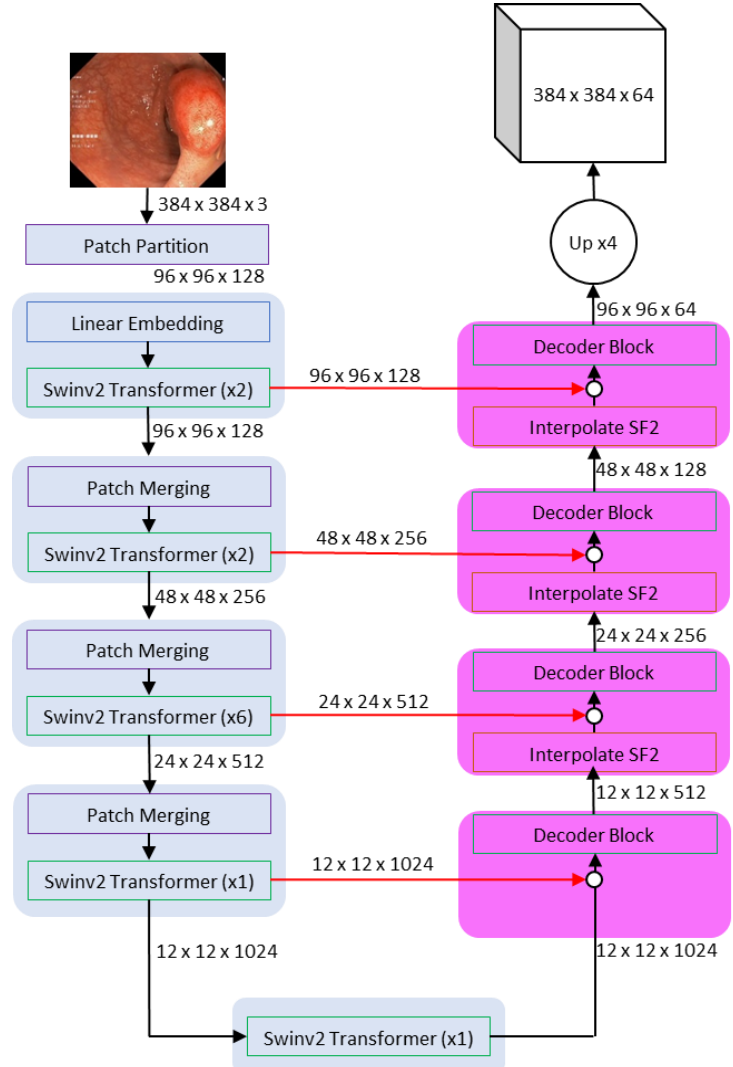

*Figure 3: SwinV2-UNET [16] architecture used as the TB of the FCB-SwinV2 Transformer. The encoder stages reduce the spatial dimensions of feature maps while increasing the number of channel dimensions. Skip connections are used to pass feature maps generated by each stage of the encoder to decoder stages. The encoder is pre-trained using ImageNet22K [17].*

Small changes to the FCB were also made compared to the original FCBFormer system. These included an increased number of channel dimensions (to match the number of channel dimensions output from the TB) and a change in the order of group normalization (GN) in the FCB residual block (RB) inspired by the residual post normalization approach of the SwinV2 Transformer. The overall architecture of the FCB-SwinV2 Transformer and RB changes are shown in Figure 4.

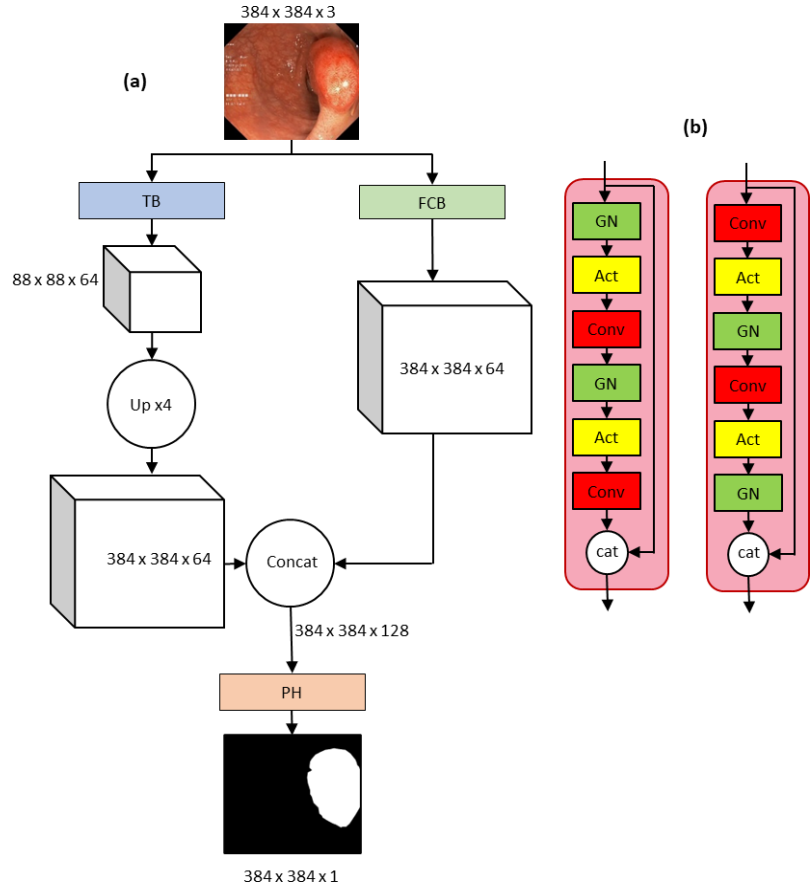

Figure 4: (a) Overall FCB-SwinV2 Transformer architecture. (b) Original RB used by the FCBFormer (left) vs the RB used by the FCB-SwinV2 Transformer (right) which features residual post normalization.

An additional system was also tested as a replacement for the TB. This additional system was based very closely on the original, fully-attention based Swin-UNET [21] with the only modifications relating to replacement of the original Swin Transformer blocks with SwinV2 Transformer blocks. Testing showed that when using this system as the TB, the overall FCB-SwinV2 Transformer segmentation performance on colonoscopy datasets was excellent (>92% mDice) but it did not beat current state-of-the-art results and so is not discussed further.

## 3 Experiments

### 3.1 Dataset Selection

The datasets Kvasir-SEG [10] and CVC-ClinicDB [11] have been used in this work to evaluate the performance of the FCB-SwinV2 Transformer. These datasets have been chosen because they are open access at the time of writing (which is not true for other popular datasets reported within literature) and because they are commonly used in colonoscopy segmentation literature to evaluate model performance. The Kvasir-SEG dataset consists of 1000 images of polyps and corresponding ground truth binary segmentation masks of varying resolutions. The CVC-ClinicDB dataset consists of 612 images of polyps and corresponding ground truth binary segmentation masks of standard resolution 384x288. After examining recent literature on deep learning polyp segmentation models which use these datasets for evaluation two important issues have been identified.

Firstly, many authors evaluating polyp segmentation models use popular dataset splitting functions (such as the train_test_split function from the scikit-learn python module) in order to create random training/validation/test data partitions (typically using an 80%/10%/10% ratio). However, for the relatively small image datasets used to evaluate colonoscopy segmentation models, minor differences in how the data is partitioned can cause noticeable performance changes. Recently this has been demonstrated for models being evaluated on the Kvasir-SEG dataset where performance changes greater than 1% were shown to occur for different data partitions [22]. Random number seeds are often used to control the data partition but the exact methodologies used to create data loaders and the specific random number seeds chosen are often not defined in enough detail. Differences in computer platforms and hidden random number state settings also exacerbate this problem.

Secondly, many authors using the CVC-ClinicDB dataset [11] create random training/validation/test data partitions but do not provide evidence to suggest they have taken steps to avoid data leakage. Data leakage is possible in the CVC-ClinicDB dataset as the 612 images are from video frames that have been taken from 29 video sequences. It is highly likely that frames from the same video sequences are present across the training, validation and test data partitions that are evaluated and reported in literature when random splits have been used.

## 3.2 Implementation Details

The FCB SwinV2 Transformer model was implemented using PyTorch. The model was trained and evaluated on images of resolution 384x384 and predicted binary segmentation maps of resolution 384x384. This resolution was used due to the availability of ImageNet [20] pre-trained SwinV2 transformer models. For each of the Kvasir-SEG and CVC-ClinicDB datasets the model was trained for 200 epochs with the loss function consisting of the sum of the Binary Cross Entropy (BCE) loss and dice loss. All training was completed using a single Nvidia 3090 GPU which necessitated a batch size of 2. The AdamW [23] optimizer was used with an initial learning rate of 1e-5. The learning rate was reduced by a factor of 0.6 when the training loss did not improve over 10 epochs. Model weights were saved each time the validation dice score surpassed the previous maximum score.

The data augmentations used in this work closely follows those employed by the authors of the original FCBFormer [9]. Geometrical data augmentations applied to the training images and masks included: vertical and horizontal flips with a probability of 0.5; scaling with a magnitude sampled uniformly from [0.5, 1.5]; shearing with an angle sampled uniformly from [-22.5°, 22.5°]; and affine transformations with rotations. (horizontal and vertical translations are sampled uniformly from [-48,48] with rotation angles being sampled uniformly from [-180°, 180°]). Color data augmentations were applied to the training images only and included: color jitter with brightness factor sampled uniformly from [0.6, 1.4], contrast factor sampled uniformly from [0.5, 1.5], saturation factor sampled uniformly from [0.75, 1.25] and hue factor sampled uniformly from [0.99, 1.01]; gaussian blur with kernel size 25x25 and standard deviation sampled uniformly from [0.001, 2.0]; and normalization of RGB images values between the interval [-1, 1] (normalization was conducted only after all other data augmentations have occurred).

Due to the issue of noticeable performance changes due to different data partitions (described in Section 3.1.), in this work we evaluate and compare performance of our model on the Kvasir-SEG [10] dataset to results reported for the fixed data partition used in [22]. This fixed data partition was created by reading Kvasir-SEG image names into a list and then sorting the names using the Python sorted function. The first 80% of the indices of the sorted list make up the training partition, the next 10% of the indices make up the validation partition, with the remaining 10% of the indices making up the test partition.

Due to the issue of data leakage being possible between partitions in the CVC-ClinicDB dataset (described in Section 3.1. and touched upon in [2]), two results for the CVC-ClinicDB are reported in this work. The first result is for a random training/validation/test data partition with an 80%/10%/10% ratio in order to compare to state-of-the-art results reported in recent literature. The second result is for a training/validation/test data partition with an 82.35%/8.82%/8.82% ratio which ensures no data leakage occurs. Frames from video sequences 4, 19 and 26 are used for validation with frames from video sequences 11, 18 and 23 used for testing.

Generalizability tests are also conducted in this work. The model trained on the Kvasir-SEG dataset (using the same 80%/10% training/validation fixed data partition) is evaluated on the full CVC-ClinicDB dataset and the model trained on the CVC-ClinicDB dataset (using the same 80%/10% training/validation random data partition) is evaluated on the full Kvasir-SEG dataset. These tests give an indication of how the models performs with respect to a different data distribution and help alleviate the identified issues described in Section 3.1. as they greatly reduce the impact of data partition changes and eliminate data from the training partition leaking into the test partition.

## 4    Evaluation

The performance of the FCB-SwinV2 Transformer was evaluated using mDice, mIoU, mPrecision and mRecall metrics (where 'm' represents the mean of the evaluation metric over the test set). Where possible evaluation metrics are compared to those attained by other high performing colonoscopy segmentation models in the literature.

The performance of the FCB-SwinV2 Transformer for the Kvasir-SEG [10] dataset are reported in Table 1. Comparisons to FCBFormer model performance using the same fixed data partition, with and without the advanced data augmentation technique named "Spatially Exclusive Pasting" (SEP) [22], are provided alongside the random data partition reported in the original FCBFormer paper [9]. In addition a comparison against other high performing models on random data partitions is given. The FCB-SwinV2 Transformer achieves the highest performance on mDice and mPrec metrics.

*Table 1: Comparison of FCB-SwinV2 Transformer model performance on the Kvasir-SEG dataset against other high performing models. The * sign indicates that the results were provided for a random data partition test set.*

| Metric | mDice | mIoU | mRec | mPrec |
|---|---|---|---|---|
| MSRF-Net* [5] | 92.17 | 89.14 | **96.66** | 91.98 |
| ESFPNet-L* [24] | 93.10 | 88.70 | - | - |
| FCB-Former* [9] | 93.85 | 89.03 | 94.59 | 94.01 |
| FCB-Former [22] | 93.91 | 89.84 | - | - |
| FCB-Former + SEP [22] | 94.11 | **90.02** | - | - |
| FCB-SwinV2 Transformer | **94.20** | 89.73 | 93.49 | **95.85** |

The performance of the FCB-SwinV2 Transformer for the CVC-ClinicDB [11] dataset are reported in Table 2. Comparisons for each model are reported for random data partitions. In addition a result for the FCB-SwinV2 Transformer for the training/validation/test data partition which ensured no data leakage (NDL) occurs is also reported. The FCB-SwinV2 Transformer achieves the highest performance across all metrics when evaluated using a random data partition. However, when no data leakage occurs the performance of the model, whilst still strong, drops significantly across all metrics.

*Table 2: Comparison of FCB-SwinV2 Transformer model performance on CVC-ClinicDB dataset against other high performing models. The * sign indicates that the results were provided for a random data partition test set.*

| Metric | mDice | mIoU | mRec | mPrec |
|---|---|---|---|---|
| MSRF-Net [5]* | 94.20 | 90.43 | 94.27 | 95.67 |
| ESFPNet-L [24]* | 94.90 | 90.70 | - | - |
| FCB-Former [9]* | 94.69 | 90.20 | 95.25 | 94.41 |
| FCB-SwinV2 Transformer* | **95.73** | **91.89** | **95.62** | **95.94** |
| FCB-SwinV2 Transformer (NDL) | 90.01 | 82.61 | 94.39 | 86.99 |

The generalizability performance of the FCB-SwinV2 Transformer is compared to that of the FCBFormer in Table 3. When trained using the Kvasir-SEG dataset and evaluated using CVC-ClinicDB as the test set, the FCB-SwinV2 Transformer achieves higher performance on the mDice and mPrec metrics. When trained using the CVC-ClinicDB dataset and evaluated using Kvasir-SEG as the test set, the FCB-SwinV2 Transformer achieves higher performance on the mDice , mIoU and mRec metrics.

*Table 3: Comparison of the generalizability performance of the FCB-SwinV2 Transformer against that of the FCBFormer.*

| Train | Kvasir-SEG | | | | CVC-ClinicDB | | | |
|---|---|---|---|---|---|---|---|---|
| Test | CVC-ClinicDB | | | | Kvasir-SEG | | | |
| Metric | mDice | mIoU | mRec | mPrec | mDice | mIoU | mRec | mPrec |
| FCB-Former [9] | 87.35 | **80.38** | **89.95** | 88.76 | 88.48 | 82.14 | 93.54 | **87.54** |
| FCB-SwinV2 Transformer | **87.84** | 80.08 | 89.40 | **89.29** | **89.61** | **82.84** | **94.46** | 87.39 |

Visual comparisons of predictions made by the FCB-SwinV2 Transformer for the Kvasir-SEG dataset when trained using the fixed data partition and when trained using the CVC-ClinicDB dataset are provided in Figure 5. Qualitative inspection of the predicted binary segmentation maps generated when trained using the fixed data partition show the strong performance of the model. Qualitative inspection of the predicted binary segmentation maps generated when trained using the CVC-ClinicDB dataset show that the model generalizes well for regular polyps but suffers a performance drop for large, irregular polyps (which are considerably different to any of the polyps within the CVC-ClinicDB dataset). Another interesting finding of the visual inspection is that some ground truth maps of the Kvasir-SEG dataset contain sharp edges (highlighted using red arrows within Figure 5) when the polyp contained within the input image appears to have smooth edges. The segmentation predictions made by the FCB-SwinV2 Transformer typically contain smoother edges. This may suggest that predictions made by the FCB-SwinV2 Transformer (and other recent state-of-the-art models) may be approaching or have surpassed the maximum achievable performance when trained and evaluated on Kvasir-SEG dataset.

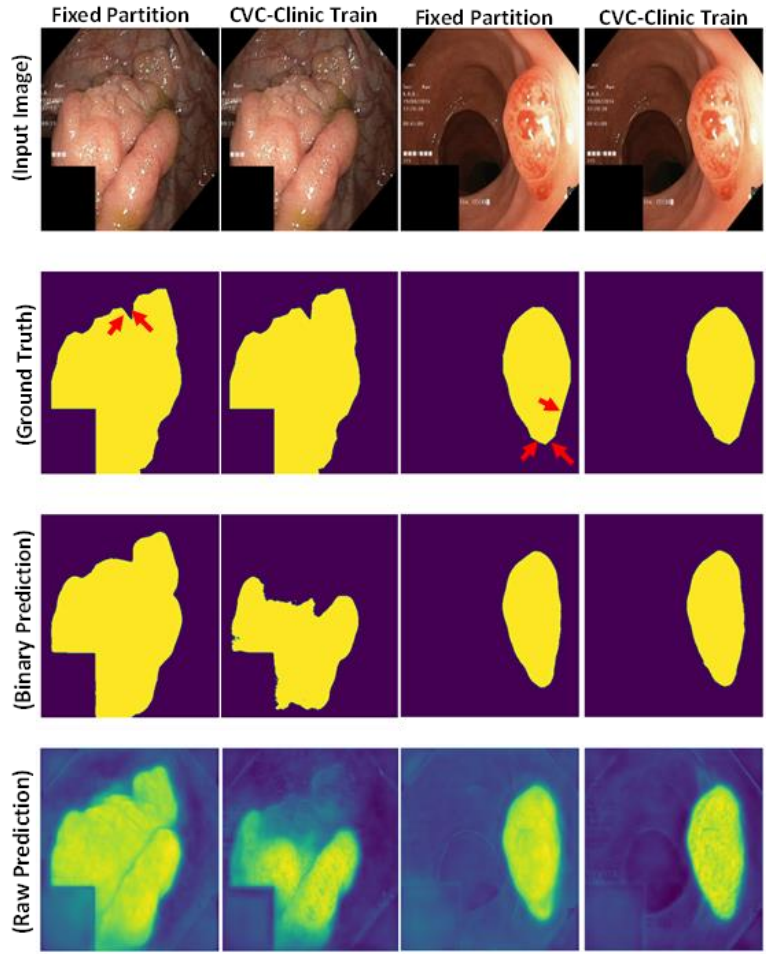

*Figure 5: Visual comparisons of predictions made by the FCB-SwinV2 Transformer for the Kvasir-SEG [10] dataset when trained using the fixed data partition and when trained using the CVC-ClinicDB [11] dataset. Red arrows highlight sharp edges found within ground truth segmentation maps which do not appear to match polyp edges within the input image. The generalizability of the model is good for regular polyps but suffers a large performance drop for certain large, irregular polyps.*

Visual comparisons of predictions made by the FCB-SwinV2 Transformer for the CVC-ClinicDB dataset when trained using the random data partition, when trained using the data partition which ensured no data leakage and when trained using the Kvasir-SEG dataset are provided in Figure 6 for a polyp with a unique shape. Qualitative inspection of the predicted binary segmentation map generated when using the random data partition shows that it matches the ground truth extremely well. When compared to the segmentation map generated ensuring no data leakage occurred we see a large drop in performance. This strongly demonstrates the artificially high performance when using a random data partition as the model has been trained and evaluated on images of the same polyp from a video sequence. The generalization performance of the model is shown to produce a significantly improved segmentation map over the one generated with no data leakage, mainly missing only the unusual upper right structure of the polyp.

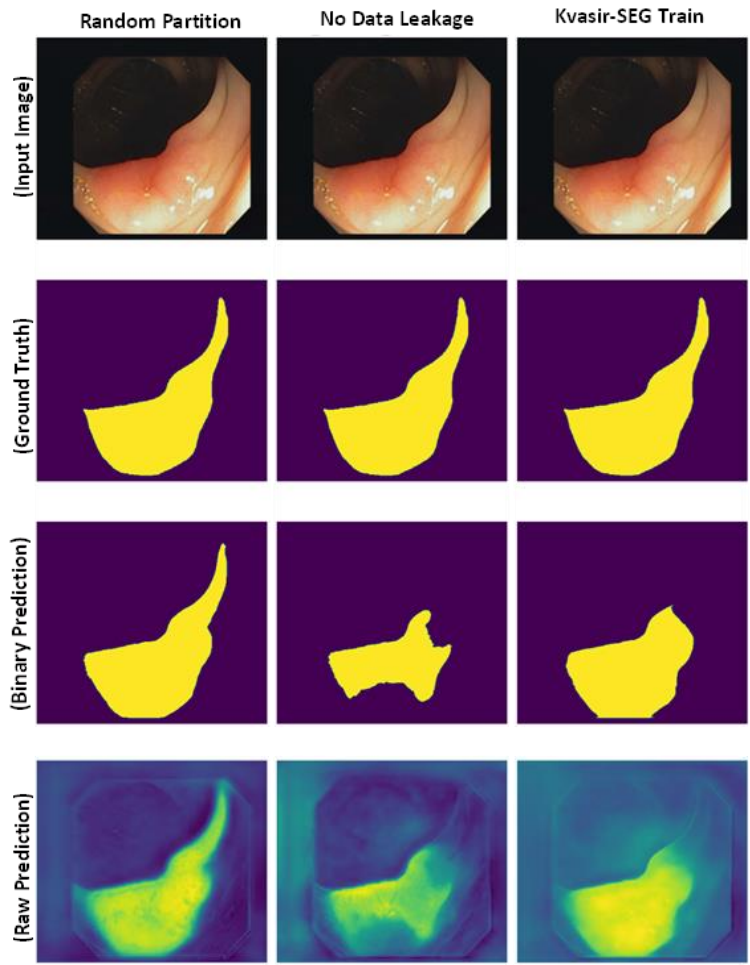

*Figure 6: Visual comparisons of predictions made by the FCB-SwinV2 Transformer for the CVC-ClinicDB [11] dataset when trained using the random data partition, when trained using the data partition which ensured no data leakage and when trained using the Kvasir-SEG [10] dataset. When using a random data partition, the model has been trained and evaluated on images of the same polyp from a video sequence resulting in artificially high performance.*

## 5    Conclusion

The FCB-SwinV2 Transformer is able to consistently achieve the highest mDice scores across all tests conducted and therefore represents new state-of-the-art performance when evaluated against this metric. Furthermore, the improved generalizability performance and improved performance when using the fixed data partition for the Kvasir-SEG dataset provides increased confidence that model comparisons are fair. The improvement in performance is hypothesized to be a result of a number of factors including: the proven ability of SwinV2 to act as a strong encoder; the residual post normalization approach within the RB of the FCB; and increased channel dimension of the FCB. The use of the SCSE module in the decoder block could also be responsible for improving performance. Testing each of these factors in different combinations or in isolation would help evaluate and pinpoint the reasons for performance improvement but would be computationally expensive to conduct.

An extremely important finding of this work is that in the future, authors evaluating performance on the CVC-ClinicDB dataset need to ensure no data leakage from video sequences occurs during the training/validation/test data partition. Images from the same video sequence are very similar and any data leakage has been strongly shown to impact performance in an artificially positive way. The colonoscopy segmentation community could benefit from expert clinicians creating standardized K-Fold cross validation data splits in the CVC-ClinicDB, Kvasir-SEG and other popular datasets reported in the literature. In addition, the qualitative examination of segmentation mask predictions, ground truth maps and images within the Kvasir-SEG dataset may suggest that predictions made by the FCB-SwinV2 Transformer (and other recent state-of-the-art models) may be approaching or have surpassed the maximum achievable performance when trained and evaluated on Kvasir-SEG dataset. These points further highlight the importance of generalizability tests.

Improvements to the FCB-SwinV2 Transformer model and training process could also be achieved. Using an advanced data augmentation technique (such as Spatially Exclusive Pasting [22] or advanced image warping) could improve the training process and increase performance on test data sets. The FCB branch could be replaced by an ImageNet pre-trained FCN model (such as [5]) to investigate if this could improve performance. There is also be the possibility of using larger (e.g. greater number of layers, channel dimensions etc.) ImageNet pre-trained SwinV2 encoders.